\documentclass[10pt]{article}
\usepackage{booktabs}
\usepackage{caption}
\usepackage{subcaption}
\usepackage[top=0.9in, bottom=0.9in, left=0.75in, right=0.75in]{geometry}

\usepackage{graphicx}
\usepackage{multirow,amssymb}
\usepackage[table,xcdraw,dvipsnames]{xcolor}
\usepackage{amsmath,amssymb}
\usepackage{multirow}
\usepackage{amsmath}
\usepackage[inline]{enumitem}
\usepackage{array,booktabs}
\usepackage{graphicx,xspace}
\usepackage{adjustbox,makecell}
\usepackage[bb=boondox]{mathalfa}
\usepackage[pagebackref,breaklinks,colorlinks]{hyperref}
\usepackage[capitalize]{cleveref}
\usepackage[accsupp]{axessibility} 
\usepackage{graphicx}
\begin{document}

\title{\vspace{-2cm}Herd Mentality in Augmentation - Not a Good Idea! \\ A Robust Multi-stage Approach towards Deepfake Detection}
\author{ Monu\thanks{Corresponding Author (\url{cs5200432@cse.iitd.ac.in})}, Rohan Raju Dhanakshirur\\
    \normalsize  Indian Institute of Technology Delhi, New Delhi, India
}
\date{}
\maketitle

\begin{abstract}
\normalsize
 \noindent The rapid increase in deepfake technology has raised significant concerns about digital media integrity. Detecting deepfakes is crucial for safeguarding digital media. However, most standard image classifiers fail to distinguish between fake and real faces. Our analysis reveals that this failure is due to the model's inability to explicitly focus on the artefacts typically in deepfakes. We propose an enhanced architecture based on the GenConViT model, which incorporates weighted loss and update augmentation techniques and includes masked eye pretraining. This proposed model improves the F1 score by 1.71\% and the accuracy by 4.34\% on the Celeb-DF v2 dataset. The source code for our model is available at
\url{https://github.com/Monu-Khicher-1/multi-stage-learning}.
\end{abstract}

\section{Introduction}

\normalsize
\textbf{Background:} Deepfakes represent manipulative media generated through advanced Deep Learning (DL) techniques, achieved by superimposing the faces of two different individuals. These manipulations encompass a variety of techniques, including entire face synthesis, identity swapping, face morphing, attribute manipulation, and expression swapping (also known as face reenactment or talking faces) \cite{cite10, cite12}. With the progression of DL methods, the realism and believability of such content have significantly increased \cite{cite11}. The emergence of deepfakes has raised public concerns due to their potential to deceive, misuse, and disseminate false information \cite{cite12, cite13}. To mitigate the spread of misleading information, researchers have introduced numerous deepfake detection techniques \cite{cite6, cite7, cite8}. While most of these approaches employ binary classifiers and demonstrate high performance in terms of accuracy and AUC, they often exhibit poor F1 scores, highlighting their limitations in effectively distinguishing between authentic and manipulated images. 

\noindent \textbf{Datasets \& Existing Models:} Deepfake Detection has been a well-explored research topic. Researchers also proposed multiple datasets for the problem because of the data-driven nature of existing computer vision techniques. Andreas Rossler \cite{cite16} proposed a deepfake videos dataset in 2019. After that, Yuezun Li \cite{cite15} also proposed a deepfake video dataset, CelebDF. CelebDF includes 590 real videos and 5,639 high-quality fake videos crafted by the improved DeepFake algorithm \cite{cite15}. We evaluated our proposed model and existing models on CelebDF. Further, we converted this video dataset into an image dataset by extracting frames at a 1fps rate. Then, we divided this image dataset into train, validation, and testing datasets (6:2:2) to evaluate and train existing models and our model.

\noindent The number of papers published in the deepfake detection domain is increasing with advancements in deepfake generative models. The publication of deepfake detection papers has increased in the past few years. Most papers are based on binary classifiers where the deepfake detection problem is treated as a general image classification problem. Broadly, the deepfake detection problem can be solved by two methods: binary classifier \& others.

\noindent \textbf{Binary Classifiers:} Deepfake Detection is treated as a binary image classification problem [17, 18] in the initial few years. Usually, they use a backbone (e.g. Xception \cite{cite19}, VGGNet \cite{cite21}) encoder to extract higher-level feature maps for input images and then use these to classify them using a binary classifier. Recently, an integrated feature extraction backbone is proposed after publishing the Attention paper \cite{cite22}. For example, Zhou et al. \cite{cite20} designed a multi-attentional face forgery detector that aggregates the texture features and high-level semantic features of multiple local parts to classify real and fake \cite{cite8, cite20}. 

\noindent \textbf{Others:}  Many attempts are made to improve the capability of the general deep fake detectors by various handcrafted techniques. One of these methods is Reconstruction Learning in unsupervised settings \cite{cite23, cite24, cite25, cite26, cite27}. It helps the model learn more about the input and reconstruct it using encoded information. Reconstruction learning is also done in the forgery detection domain \cite{cite8, cite28}. Junyi Cao \cite{cite8} has proposed a reconstruction encoder which learns to regenerate real images and is later used as a classifier for fake and real. In some supervised methods, a reconstructor learns latent data distribution \cite{cite9}. Shichao Dong \cite{cite7} proposed an architecture comprising an artefact detection module and a feature map generation module for forgery detection. They claimed identity leakage in existing binary classifier models and utilized artefact detection architecture for deepfake detection \cite{cite7}. 

\noindent \textbf{Problems with the Existing Approaches:}  Current state-of-the-art image classifiers often fail to accurately capture the feature details of deepfakes, resulting in an inability to distinguish between fake and real images. Many existing methods also rely on these image classifiers as their backbone. GenConViT \cite{cite9} employs both feature extraction from image classifiers and data distribution extraction from autoencoders, yet it still fails in approximately 6.67\% of cases. This failure is primarily due to the inappropriate selection of data augmentation techniques for the training dataset. They \cite{cite9} use standard augmentation techniques such as Gaussian noise, random brightness contrast, and sharpening, which generate fake images and compromise the ideal conditions for deepfake detection. Our analysis of the learning patterns of deep neural networks in the context of deepfake detection reveals that the model primarily focuses on human eyes as the distinguishing feature. Although human eyes are a prominent feature, a model that concentrates on a limited feature set is highly susceptible to overfitting. Additionally, we observe that every deepfake detection dataset exhibits class imbalance.

\noindent \textbf{Our Strategy:}
\begin{enumerate*}[label=(\arabic*)]

\item The best-performing state-of-the-art model \cite{cite9} employs standard augmentation techniques such as Gaussian noise, random brightness contrast, and sharpening. These techniques generate fake images and disrupt the ideal conditions for deepfake detection. We revert to basic augmentations, such as rotation and flipping only. This minor adjustment results in an 8.21\% increase in the F1 score and a 3.85\% increase in accuracy for the CelebDF V-2 \cite{cite6} dataset.

\item The model \cite{cite9} predominantly focuses on human eyes as the distinguishing feature. We propose a hardness-inspired curriculum in which we pre-train the model on a dataset with synthetically masked eyes. This pre-training enables the model to learn other features. The pre-trained model is then further trained on the actual dataset. This modification results in an additional ~1.0\% improvement in the F1 score.

\item We address class imbalance using a weighted loss function. This adjustment further enhances the base model's performance by 1.64\% in the F1 score.

\end{enumerate*}

\noindent \textbf{Contribution:}  We propose a multi-step approach for better deepfake detection. Specifically:
\begin{enumerate*}[label=(\arabic*)] 
\item We review top deepfake detection model \cite{cite9} and find that the wrong data augmentation techniques cause them to fail. By using simpler techniques like rotation and flipping, we improve the F1 score by 8.21\% and accuracy by 3.85\% on the CelebDF V-2 dataset.
\item We notice that models focus too much on human eyes, leading to overfitting. To fix this, we pre-train a model on a dataset with masked eyes to help it learn other features. This results in a 1.0\% improvement in the F1 score.
\item We address class imbalance in deepfake detection datasets by using a weighted loss function, which further improves the F1 score by 1.64%.
\item Overall, we improved the F1 score by 1.71\% (for CADDM, the F1 score is 93.5\%, and accuracy by 5.03\%(for GenConViT \cite{cite9}, the accuracy is 93.33\%). 
\end{enumerate*}
% \section{Related Work}

% \newgeometry{bottom=2in}

\section{Proposed Architecture}

\subsection{Backbone Architecture:} 
\begin{figure}[h!]
    \centering
    \begin{subfigure}[b]{0.158\textwidth}
        \includegraphics[width=\textwidth]{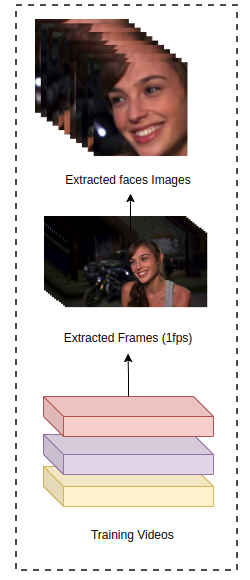}
        \caption{}
         \label{fig:model1a}
    \end{subfigure}
    \begin{subfigure}[b]{0.805\textwidth}
        \includegraphics[width=\textwidth]{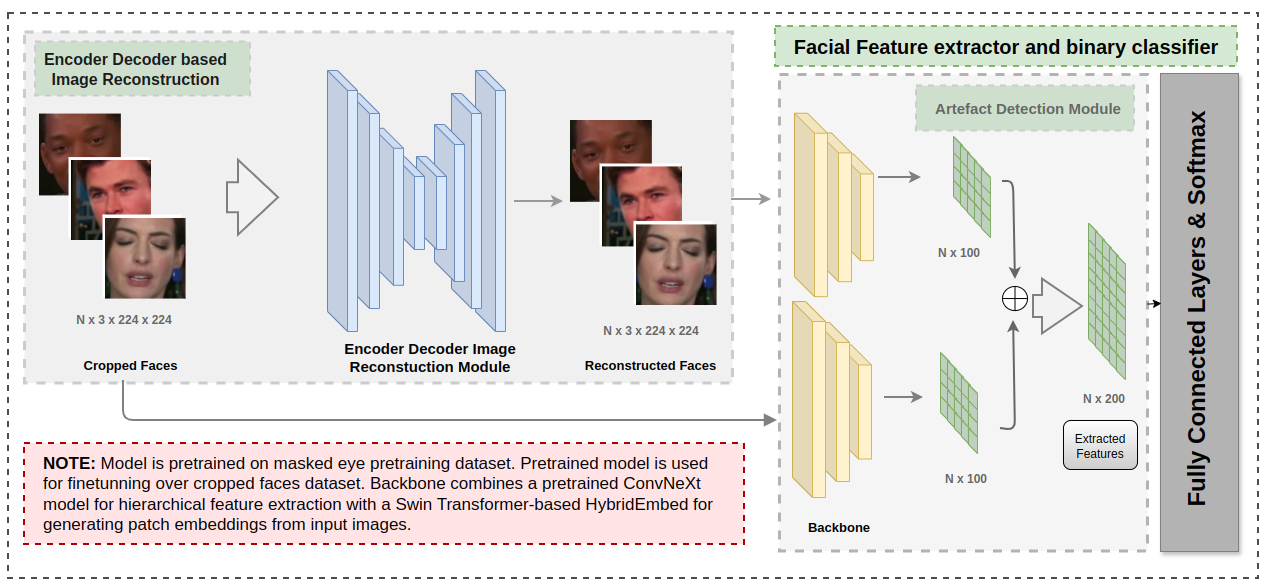}
        \caption{}
         \label{fig:model1b}
    \end{subfigure}
    \caption{ (a) State of the art data preprocessing as used in \cite{cite9} (b)  Baseline \cite{cite9} model architecture }
    \label{fig:model1}
\end{figure}

We utilized GenConViT \cite{cite9} as backbone architecture for our model. As illustrated in Figure~\ref{fig:model1b}, it uses an Encoder-Decoder for reconstructing images and learns latent data distribution through this Autoencoder. Backbone of GenConViT consists of the SwinTansformer \& ConvNeXt layer \cite{cite9}.

 The ConvNeXt and Swin transformer models form a hybrid ConvNeXt-Swin, represented as the backbone in Figure~\ref{fig:model1b}. The ConvNeXt model acts as the backbone of the hybrid model, using a CNN to extract features from the input frames. The Swin Transformer, with its hierarchical feature representation and attention mechanism, further extracts the global and local features of the input images \cite{cite9}. Backbone (ConvNeXt-Swin) extracts out visual features of the input image. Later, these features are used by the classifier.  The model takes N images as input; the Encoder will generate N x 256 x 7 x7 feature space for N x 3 x 224 x 224, which will be input to the decoder to reconstruct the original image. Backbone (ConvNeXt-Swin) will generate an N x 100 feature map for input and reconstructed images. These feature maps are concatenated and used as input in the binary classifier to make final predictions.

\subsection{Problems with GenConViT:} The model predicts various outputs using the feature map generated by the backbone. We observed some problems during our analyses:
\begin{enumerate}
\item Our analysis reveals that real faces are misclassified more than fake faces. Out of 13,543 fake images, 863 images are misclassified by backbone architecture, i.e. 6.37\% of fake images are misclassified. On the other hand, 135 out of 1,415 real images are misclassified, i.e., 9.5\% are misclassified.
\item GenConViT has used augmentations, such as GaussNoise, RandomBrightnessContrast, Sharpening, etc., during training, which involves manipulating the original dataset. Because if noise is added to a real image, it will become fake. Wrong Augmentation techniques are used.
\item GradCAM analyses for ConNeXt's last convolution layers reveal that the model focuses more on eyes and background than other facial details. It bounds the model to classify according to only eye artefacts.
\end{enumerate}

\subsection{Modified Data Augmentation(NA \& RF):} 

Standard augmentation techniques such as Gaussian noise, random brightness contrast, and sharpening generate fake images and disrupt the ideal conditions for deepfake detection. Augmentation can be used for this task, which doesn’t introduce any noise and just rotates the image. So, we revert to basic augmentations, such as rotation and flipping only. This minor adjustment results in an 8.21\% increase in the F1 score and a 3.85\% increase in accuracy for the CelebDF V-2 \cite{cite6} dataset Table~\ref{tab:model_performance2}. We proposed 2 methodology for it: (1) No Augmentation (NA) \& (2) Random Flips (RF) (flipping-based augmentation)

\subsection{Hardness-inspired Curriculum Learning (MEP):} 
\begin{figure}[h!]
    \centering
    \begin{subfigure}[b]{0.60\textwidth}
        \includegraphics[width=\textwidth]{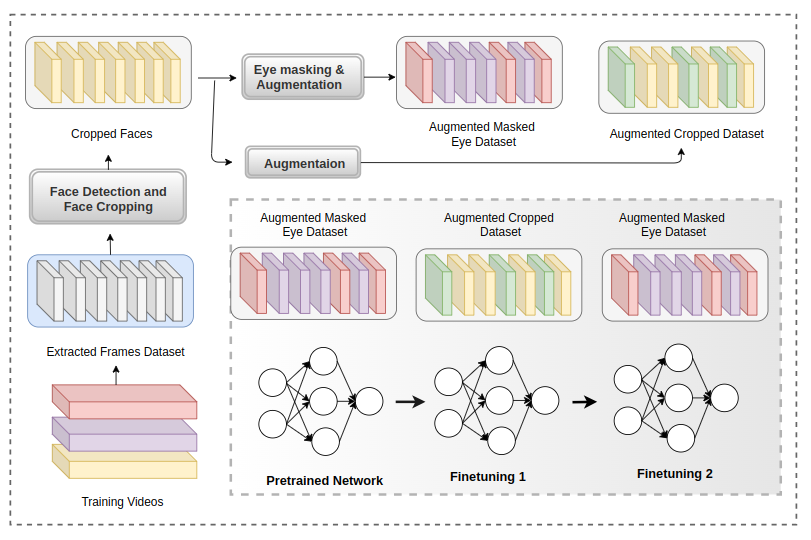}
        \caption{multistage training architecture}
        \label{fig:model2a}
    \end{subfigure}
    \begin{subfigure}[b]{0.38\textwidth}
        \includegraphics[width=\textwidth]{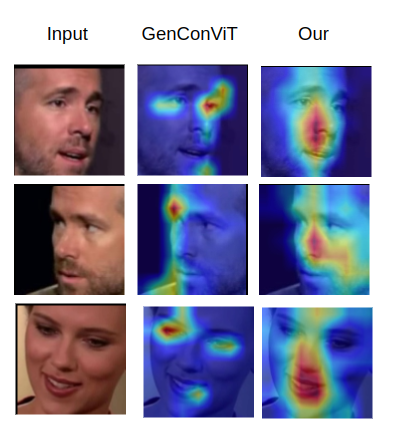}
        \caption{Grad-cam attention map }
        \label{fig:model2b}
    \end{subfigure}
    \caption{Architecture analyses with GradCAM}
    \label{fig:model2}
\end{figure}

\textbf{Multi-stage Training:} Proposed new Multi-stage training methodology(Figure ~\ref{fig:model2}) for better training of model weights such that it makes predictions based on facial features other than eyes also. For this, we trained the model Augmented Masked Eye Training dataset and then fine-tuned on Augmented Cropped Faces Dataset \& then finetuned on Augmented Masked Eye Training dataset respectively. More details: Figure ~\ref{fig:model2}.

\subsection{Weighted Loss Function (WL):} The dataset exhibits class imbalance, with fake images being nearly ten times more prevalent than real images in the test dataset. GenConViT \cite{cite9} misclassifies real faces more frequently than fake faces. To address this issue, we introduced a weighted loss function:
\[L_{fake} = w*L_{real}\].
After tuning the hyperparameter $w$, we found the optimal value to be 
$w=1.85$. This adjustment improved the F1 score by 4.46\% for the base model. Ablation analyses with other approaches are shown in Table~\ref{tab:model_performance1}.

\section{Results and Discussion}
\subsection{Dataset under consideration:}
We demonstrated the performance of the proposed methodology on CelebDF-v2 \cite{cite15}.  CelebDF includes 590 real videos and 5,639 high-quality fake videos crafted by the improved DeepFake algorithm \cite{cite15}. This video dataset is preprocessed to extract frames from videos (1fps). This image dataset is split into training, validation and testing (6:2:2). Then, the training dataset is balanced. After preprocessing and all other steps, the training dataset contains 4000 real images and 4000 fake images. Test Dataset is imbalanced and contains 1,415 real and 13,543 fake images. Further, validation have 1,137 real and 10,833 fake images. Some models used this dataset.

% \subsection{Implementation Details:}
\subsection{Evaluation Metrics:}

We employed several evaluation metrics, including the F1 score, accuracy, and area under the curve (AUC), to analyze the results of different methodologies and models. Although accuracy is commonly used for comparing different models, it is not an appropriate parameter when dealing with imbalanced test datasets. Given that most deepfake datasets exhibit significant class imbalance, relying on accuracy alone can be misleading and fails to provide a comprehensive understanding of a model's performance.

\noindent In scenarios with imbalanced datasets, the F1 score is more suitable as it considers both precision and recall, offering a balanced measure of a model's performance. Therefore, we utilized the F1 score for evaluating deepfake detection models. Additionally, we also employed the AUC, specifically the area under the receiver operating characteristic curve (AuROC), for further analysis. 

\subsection{Quantitative Results on CelebDF:}

The results of the proposed method \& comparison with SOTA deepfake detection models are shown in Table~\ref{tab:model_performance2}.  It can be observed that the proposed method outperforms the best-performing SOTA method by 1.71\%.  Table~\ref{tab:model_performance1} compares the performance of our base model with other models using different combinations of techniques. Table~\ref{tab:model_performance2} shows the performance of our approach compared with existing state-of-the-art models.

\begin{table}[ht]
  \centering
  \begin{minipage}{.45\linewidth}
    \centering
    \begin{tabular}{lccc}
        \toprule
        \textbf{Model} & \textbf{Acc} & \textbf{F1} & \textbf{AUC} \\
        \midrule
        ResNet-50      & 0.9200       & 0.5880      & 0.9190       \\
        SENet          & 0.9200       & 0.6480      & 0.9450       \\
        EfficientNet-b7& 0.9480       & 0.7080      & 0.9490       \\
        BiT            & 0.9160       & 0.5420      & 0.8830       \\
        ViT            & 0.9200       & 0.6062      & 0.9300       \\
        RECCE \cite{cite8}        & 0.8617       & 0.8483      & 0.9074       \\
        CADDM \cite{cite7}         & 0.7900       & 0.9350      & 0.9970       \\
        GenConViT \cite{cite9}      & 0.9333       & 0.8408      & 0.9770       \\
        Ours           & \textbf{0.9836} & \textbf{0.9521} & \textbf{0.9952} \\
        \bottomrule
    \end{tabular}
    \caption{Performance comparison with state-of-the-art models}
    \label{tab:model_performance2}
  \end{minipage}%
  \hfill
  \begin{minipage}{.45\linewidth}
    \centering
    \begin{tabular}{lccc}
        \toprule
        \textbf{Model} & \textbf{Acc} & \textbf{F1} & \textbf{AUC} \\
        \midrule
        Base model     & 0.9333       & 0.8408      & 0.9770       \\
        WL             & 0.9560       & 0.8854      & 0.9560       \\
        NA             & 0.9718       & 0.9229      & 0.9939       \\
        RST            & 0.9372       & 0.8498      & 0.9824       \\
        MEP            & 0.9646       & 0.9067      & 0.9913       \\
        WL + NA        & 0.9786       & 0.9394      & 0.9939       \\
        WL + NA + RST  & 0.9692       & 0.9170      & 0.9922       \\
        WL + NA + MEP  & 0.9769       & 0.9366      & 0.9947       \\
        NA + MEP       & 0.9777       & 0.9334      & 0.9899       \\
        WL + RF + MEP  & 0.9836       & \textbf{0.9521} & \textbf{0.9952} \\
        \bottomrule
    \end{tabular}
    \caption{Performance metrics of different models with combinations of techniques (Ablation Analyses)}
    \label{tab:model_performance1}
  \end{minipage}
\end{table}

\subsection{Ablation Analysis:}
We proposed various methodologies to enhance GenConViT \cite{cite9}, addressing several network issues. During our analyses, we identified multiple problems within the model and proposed corresponding solutions:
\begin{enumerate*}[label=(\arabic*)] 
\item \textbf{Class Imbalance:} A common issue in most deepfake detectors. We found that implementing a Weighted Loss (WL) function effectively mitigates this problem.
\item \textbf{Ineffective Augmentation:} Standard augmentations, such as Gaussian noise and random brightness contrast, were counterproductive. We discovered that no augmentation (NA) was more effective than these standard methods. Additionally, we proposed a Random Flip (RF) augmentation, which involves flipping the images.
\item \textbf{Masked Eye Pretraining (MEP):} This technique is used to enhance the model's ability to make predictions based on facial features other than the eyes.
\item  \textbf{Model Complexity:} We attempted to simplify the model by removing the Swift Transformer (RST) from the backbone. However, this did not improve performance; in some misclassified cases, information loss was observed in higher-level layers, leading to the decision to retain those layers.
\end{enumerate*}

RST method didn't work with other methods. Further, RF gives better results as compared to NA. Finally, we get the best accuracy of 98.36\% and F1 score of 95.21\%.

\noindent We experimented with different combinations of the proposed solution methodology. The results are in Table~\ref{tab:model_performance1}

% \begin{table}[h]
%     \centering
%     \caption{}
%     \label{}
    
% \end{table}

% \begin{table}[h]
%     \centering
%     \caption{}
%     \label{}
    
% \end{table}

\section{Conclusion}
In this paper, we proposed methodologies to improve GenConvNet. Towards this, first, we proposed a weighted loss concept for deepfake models \& augmentations suitable for deepfakes. Then, we proposed a multi-stage training architecture for fixing architecture ablations. Overall,  the proposed method outperforms the best-performing SOTA method by 1.71\%  in the publicly available dataset Celeb-DF.

\section*{Acknowledgment}
We acknowledge Prof. Prem Kumar Kalra for his valuable guidance and contributions to this work.
% \newpage
\bibliographystyle{unsrt}
\bibliography{template}
\end{document}